\documentclass[letterpaper, 10pt, conference]{ieeeconf}      % Use this line for a4 paper

\IEEEoverridecommandlockouts                              % This command is only needed if 
\usepackage{cite}                                           % you want to use the \thanks command
\usepackage{url}
\usepackage{textcomp}
\usepackage{gensymb}
\usepackage{graphicx}
\graphicspath{ {./figures/} }
\usepackage{multicol}
\usepackage{algorithm}
\usepackage{algpseudocode}
\usepackage{amsmath}

\usepackage{comment}
\usepackage{caption}
\usepackage{subcaption}
\usepackage[utf8]{inputenc}
\usepackage{booktabs}

\begin{document}

\title{A Generative Partially Specified Finite State Machine Approach to Complex Behaviour Planning}

\author{Kalana Ratnayake$^{1}$, Michael Pritchard$^{2}$, David Hinwood$^{3}$, Maleen Jayasuriya$^{4}$, Damith Herath$^{5}$% <-this % stops a space
\thanks{$^{1}$Kalana Ratnayake is with Collaborative Robotics Laboratory, Faculty of Science and Technology, University of Canberra, ACT, Australia {\tt\small Kalana.Ratnayakemudiyanselage@canberra.edu.au}}%
\thanks{$^{2}$Michael Pritchard is with Collaborative Robotics Laboratory, Faculty of Science and Technology, University of Canberra, ACT, Australia {\tt\small Michael.Pritchard@canberra.edu.au}}%
\thanks{$^{3}$David Hinwood is with IA-Cobotics Lab, School of Engineering, RMIT, Melbourne, Australia {\tt\small David.Hinwood@rmit.edu.au}}%
\thanks{$^{4}$Maleen Jayasuriya is with Collaborative Robotics Laboratory, Faculty of Science and Technology, University of Canberra, ACT, Australia {\tt\small Maleen.Jayasuriya@canberra.edu.au}}%
\thanks{$^{5}$Damith Herath is with Collaborative Robotics Laboratory, Faculty of Science and Technology, University of Canberra, ACT, Australia {\tt\small Damith.Herath@canberra.edu.au}}%
}

\maketitle

\newcommand\blfootnote[1]{%
  \begingroup
  \renewcommand\thefootnote{\fnsymbol{footnote}}%
  \renewcommand\thempfootnote{\fnsymbol{mpfootnote}}%
  \footnotetext[0]{#1}%
  \endgroup
}

% Place this right before \maketitle
\blfootnote{© 2026 IEEE. Personal use of this material is permitted. Permission from IEEE must be obtained for all other uses, in any current or future media, including reprinting/republishing this material for advertising or promotional purposes, creating new collective works, for resale or redistribution to servers or lists, or reuse of any copyrighted component of this work in other works. This work has been accepted for publication in the 2026 IEEE/RSJ International Conference on Intelligent Robots and Systems (IROS). The final published version will be available via IEEE Xplore.}

\begin{abstract}
Autonomous robots operating in dynamic environments require behaviour planning systems that combine reactivity, interpretability, and adaptability. While Large Language Models (LLMs) have been successfully integrated with Behaviour Trees (BTs) for dynamic replanning, Finite State Machines (FSMs)—despite their widespread adoption and computational efficiency—remain unexplored for generative approaches. We propose a Generative Partially Specified Finite State Machine (GPSFSM) neurosymbolic architecture that utilises the symbolic and semantic structure of FSMs, including states and event-triggered transitions, to implement Behaviour Planning. This paper introduces the first GPSFSM framework for robotics, featuring \textit{Fabric}, an FSM engine that parses, validates, and executes behaviour plans that contain Sequential, Recovery, Parallel-Any, and Parallel-All control structures. We extend the Capabilities2 package in ROS2 with an asynchronous event system for behaviour chaining and runtime parameter injection for configurable execution, addressing the ad-hoc function representations that limit current generative systems. PromptTools provides a unified ROS 2 interface to local and cloud LLMs, with prompt buffering, enabling dynamic asynchronous composition of task and context information. Together, these components enable standardised semantic capability descriptions for robot-agnostic development. Experimental evaluation on navigation tasks demonstrates that our GPSFSM approach achieves consistently higher plan-generation success rates than the state-of-the-art BTGenBot system, particularly excelling in zero-shot scenarios where BTs typically struggle, while maintaining comparable or lower planning latency to frontier LLMs. We also demonstrate experimentally that our system can generate complex behaviour. We release an open-source ROS2 stack that makes generative FSM planning practical and reproducible for robotic systems.
\end{abstract}

\begin{keywords}
behaviour planning, ROS2, generative behaviour planning, generative finite state machine, behaviour-based robot control, neurosymbolic architecture
\end{keywords}

%%%%%%%%%%%%%%%%%%%%%%%%%%%%%%%%%%%%%%%%%%%%%%%%%%%%%%%%%%
%%%% Introduction
%%%%%%%%%%%%%%%%%%%%%%%%%%%%%%%%%%%%%%%%%%%%%%%%%%%%%%%%%%

\section{Introduction}

% --- Preface

Autonomous robots operating in dynamic, human-centric environments require sophisticated behaviour planning capabilities that extend beyond simple reactive control. Long-horizon behaviour planning—sequencing and coordinating multiple actions over extended periods while adapting to environmental changes—is essential for robots transitioning from laboratories to real-world applications, particularly in human-robot interaction (HRI) scenarios requiring dynamic adaptation under uncertainty. Currently, behaviour-based control systems—particularly Behaviour Trees (BTs) and Finite State Machines (FSMs)—remain the most widely adopted approach for composing complex behaviour.

While recent advances in deep reinforcement learning \cite{chen_crowd-robot_2019, liu_decentralized_2021} and vision-language-action models such as RT-1 \cite{brohan_rt-1_2023} and RT-2 \cite{brohan_rt-2_2023} show promise, these behaviour-based systems remain vital. First, behaviour-based systems provide transparent, interpretable execution plans, which are essential for debugging and safety verification—critical requirements in human-shared environments. Debugging and safety verification \cite{zhang25safevla} for VLA remain in early stages compared to this. Second, they offer deterministic execution guarantees with additional fallback states (FSMs) and Recovery Nodes (BTs), which learning-based systems cannot yet reliably provide \cite{neary2025}, resulting in confident failure. Additionally, by properly linking behaviours and adding behaviour-plan regeneration as a fallback or recovery option for unaccounted-for environmental stimuli, the behaviour system can provide deterministic execution even in an unpredictable environment. Third, they enable modular composition across robot subsystems, facilitating code reuse and system integration, unlike VLAs, which require fine-tuning or retraining for each robot and task \cite{neary2025}. Thus, even as learning-based approaches mature, behaviour-based control systems will remain necessary to interconnect subsystems and ensure reliable system-wide integration. These characteristics make BTs and FSMs indispensable for orchestrating complex behaviours where reliability and predictability are paramount.

When concerned with long-horizon tasks, in unpredictable environments shared with humans, BTs, FSMs, and other non-generative behaviour-planning systems share a fundamental limitation: they rely on hard-coded, human-designed behaviour sequences. This is a limitation because, to match the environment's unpredictability, the behaviour sequences must account for all possible conditions. This has motivated research into methods such as synthesising/evolving BTs to adapt to environmental changes using Reinforcement Learning (RL), Evolution-inspired learning(EL), Case-based reasoning, and Learning from demonstration \cite{IOVINO2022104096, 9562088}. The most recent addition to this is the use of generative approaches that leverage Large Language Models (LLMs) to dynamically synthesise behaviour plans \cite{ahn_as_2022, rana_sayplan_2023}. Several generative BT systems have emerged, including LLM-Brain \cite{lykov_llm-brain_2023}, BTGenBot \cite{izzo_btgenbot_2024}, and LLM-BT \cite{zhou_llm-bt_2024}, all of which exploit BehaviorTree.CPP's \cite{colledanchise2021implementation} XML-based execution plans for natural integration with LLMs. The difference between early methods and the new LLM-based approaches is that LLMs acquire general knowledge and common sense during training, which can be beneficial for generating BTs.

Furthermore, a critical challenge for all generative behaviour systems is the lack of standardised semantic representations of robot capabilities. While the original Capabilities package for ROS \cite{woodall_ros_2014} introduced YAML-based mappings between ROS primitives and natural-language descriptions, most generative systems still rely on ad-hoc behaviour descriptions embedded in prompts. This forces each implementation to define its own representation, limiting portability and hindering robot-agnostic development. Effective generative planning requires both textual execution plans and standardised semantic descriptions of capabilities and their interconnections. Capabilities2 \cite{pritchard_capabilities2_2025} addresses this issue for ROS2 interfaces. Building on its predecessor \cite{woodall_ros_2014}, it offers a higher abstraction layer that could enable robot-agnostic generative systems—though current implementations rarely leverage such standardised representations.

To address these gaps, we present a novel Generative Partially Specified Finite State Machine (GPSFSM) framework - Fabric, that aims to generalise the benefits of FSMs and BTs while reducing the complexity of the intermediate representation of the resulting Directed Graph. GPSFSM integrates LLM-based dynamic planning with graph-based behaviour chaining and provides a standardised approach to semantic capability representations. Our contributions are:

\begin{enumerate}
    \item A Novel Generative Partially Specified Finite State Machine approach, specifically designed for robotics applications, that accounts for the inherent epistemic uncertainty of long-horizon tasks as an alternative to generative BTs. To the best of our knowledge, this is the first systematic approach to combining FSMs with LLMs to generate dynamic behaviour in robotic systems.
    
    \item A Comprehensive Open-Source ROS2 Software Stack that enables prompting, generation, validation and execution of generative FSMs. Our framework provides standardised yet flexible interfaces for describing robot capabilities, enabling rapid deployment across any ROS-based robot platform.
    
    \item Comprehensive experimental comparisons with the state-of-the-art generative BT framework BTGenBot \cite{izzo_btgenbot_2024} across generation success rates and latency, demonstrating that GPSFSMs match or exceed generative BTs, along with a set of demonstrations showcasing generative capabilities for complex behaviour planning.
\end{enumerate}

The remainder of this paper is organised as follows: Section II reviews related work; Section III presents our methodology, including the \textit{Fabric} system architecture; Section IV details the experimental evaluation and results; and Section V discusses implications and future work.

%%%%%%%%%%%%%%%%%%%%%%%%%%%%%%%%%%%%%%%%%%%%%%%%%%%%%%%%%%
%%%% Literature review
%%%%%%%%%%%%%%%%%%%%%%%%%%%%%%%%%%%%%%%%%%%%%%%%%%%%%%%%%%

\section{Related Work}

\subsection{Behaviour Composition with BTs and FSMs}

Behaviour Trees organise robot actions hierarchically in a tree structure that continuously "ticks". Each node returns one of three states (RUNNING, SUCCESS, or FAILURE), enabling reactive behaviour through continuous environmental monitoring \cite{iovino_comparison_2024}. The BehaviorTree.CPP library has become the de facto standard for BTs in robotics (exemplified by their widespread successful adoption in ROS2's Nav2 stack for complex navigation and recovery behaviours), leveraging XML-based execution plans for design verification and runtime execution. BTs have three dominant structural characteristics. They have a root node, they traverse the graph on a limited set of transition conditions, and child nodes have exactly one parent node.

Finite State Machines offer an alternative paradigm where behaviours are represented as states with explicit event-triggered transitions. Systems like FlexBe \cite{zutell_ros_2022}, YASMIN \cite{gonzalez-santamarta_yasmin_2023}, and SMACC2 provide robust FSM implementations for robotics applications. Looking at these FSM systems, most, if not all, define the states and transitions directly in the code rather than through an intermediate textual representation. This can also be considered evidence of the modularity and reusability of BTs, as intermediate textual representations (Such as XML) enable the system to take on multiple forms without writing additional code.  While FSMs traditionally require more states and transitions to achieve reactivity comparable to BTs, they offer simpler representations and computational efficiency through event-driven execution, thereby avoiding BTs' continuous-ticking overhead \cite{iovino_comparison_2024}. Further, FSMs offer a general graph description which subsumes the special case BT. Fundamentally, BTs are constrained by two rules for their graphical construction. The existence of a root node and one-to-many parent-to-child node relationships. FSMs do not impose these rules on the construction of the graph.

\subsection{LLMs based behaviour generation via code}

Early attempts at LLM-based behaviour planning focused on direct code generation. Code-as-Policies \cite{liang_code_2023} proposed generating Python-based policies using an LLM and executing them to control the robot, while Palm-SayCan \cite{ahn_as_2022} pioneered this approach by using affordance functions to select an action from a set of supported actions. However, their Python-based implementation lacks the modularity and verifiability that behaviour-based systems provide.

SayPlan \cite{rana_sayplan_2023} advanced this by integrating LLMs with 3D Scene Graphs for spatial reasoning. While demonstrating sophisticated environmental understanding, the approach generates sequential code rather than leveraging established behaviour representations. BOSS \cite{zhang_bootstrap_2023} explores skill chaining through reinforcement learning primitives orchestrated by LLMs, but again bypasses formal behaviour representations.

These approaches highlight that while LLMs can generate executable code, utilising established symbolic systems offers superior modularity, verifiability, and standardisation \cite{ahn_as_2022}, motivating the shift toward generative behaviour trees.

\subsection{LLMs based BT Generation}

The emergence of XML-based execution plans in BehaviorTree.CPP enabled generative BT research. LLM-Brain \cite{lykov_llm-brain_2023} and BTGenBot \cite{izzo_btgenbot_2024} fine-tune 7B parameter models for local BT generation. LLM-Brain generates synthetic training data using text-davinci-003, creating datasets of up to 8500 BTs. BTGenBot uses 600 human-designed BTs and generates descriptions post hoc, though this risks misalignment between descriptions and implementations.

Cloud-based approaches leverage larger models: LLM-BT \cite{zhou_llm-bt_2024} focuses on dynamic BT modification, identifying failure nodes and grafting corrective sub-trees. BETR-XP-LLM \cite{styrud_automatic_2024} extends this with failure resolution by incorporating error messages. LLM-as-BT-Planner \cite{ao_llm-as-bt-planner_2025} provides complete generation-to-execution pipelines with learning-based improvements. VLM-BT \cite{wake_vlm-driven_2025} introduces visual conditions, expanding BT expressiveness but increasing semantic complexity.

Critically, these systems universally report challenges with zero-shot generation, requiring either fine-tuning, extensive prompt engineering, or multiple examples to achieve reliable results. This suggests that BTs' hierarchical structure and ticking semantics present inherent challenges for LLM understanding. Only BTGenBot \cite{izzo_btgenbot_2024} provides ROS 2-compatible implementations, thereby limiting reproducibility.

\subsection{FSM Associated with LLMs}

Despite FSMs' potential advantages for LLM integration—simpler state semantics, explicit transitions, and natural language alignment—generative FSM research remains sparse. The primary barrier has been technical: existing FSM libraries generate code-based definitions rather than intermediate textual representations amenable to LLM generation.

Limited exploration exists: State Machine Serialisation Language (SMSL) \cite{mu_look_2025} generates textual FSM representations for imitation learning environments. Outside robotics, ProtocolGPT \cite{wei_unleashing_2025} infers protocol state machines from implementations, while ChatFSM \cite{gan_can_2024} enables natural-language FSM modification via RAG-enhanced code generation. However, none address robotic behaviour planning through generative FSMs.

This gap is striking given FSMs' semantic advantages. Where BTs require understanding tree traversal and state propagation, FSMs model behaviours as discrete states with clear transitions—thereby being conceptually closer to natural-language descriptions. Given the absence of standardised semantic capability descriptions and the reliance of current systems on ad-hoc representations, this presents a clear opportunity to advance generative behaviour planning through FSM-based approaches.

%%%%%%%%%%%%%%%%%%%%%%%%%%%%%%%%%%%%%%%%%%%%%%%%%%%%%%%%%%
%%%% Methodology
%%%%%%%%%%%%%%%%%%%%%%%%%%%%%%%%%%%%%%%%%%%%%%%%%%%%%%%%%%

\section{Methodology}

One advantage of BTs for LLM-based generation is their ability to represent the Behaviour Sequence textually, independent of the BT implementation. When used with robotic systems, this enables reconfiguration of system behaviour without modifying code. In contrast, current FSM tools define behaviours through code-based states and transitions and require regeneration or recompilation to introduce changes.

Based on existing systems and the literature, we identify the following as the minimum requirements for generative behaviour planning using BTs, FSMs, or other approaches.

\begin{enumerate}
    \item A textual description of the robot's functionalities.
    \item A textual representation of the behaviour composition based on the robot's functionalities.
    \item Functionality to coordinate with LLMs to generate the textual representation.
    \item Functionality to execute the behaviours as described in the textual representation.
\end{enumerate}

In addition to the aforementioned requirements, a key design goal was to dynamically manage the execution of behaviour. Rather than the conventional approach of loading all behaviours into memory at startup, our system allows individual behaviours to be loaded and unloaded at runtime. This dynamic architecture directly supports our secondary design goals:

\begin{itemize}
    \item \textit{Optimise resource allocation}: Reduce overall computational overhead, allowing active systems to utilise more resources.
    \item \textit{Extend functional duration}: Conserve the robot's energy by powering down unused components.
    \item \textit{Improve fault tolerance}: Minimise the probability of system failure by keeping inactive components offline.
    \item \textit{Handle epistemic uncertainty}: Adapt to long-horizon tasks where the exact sequence of required behaviours cannot be anticipated in advance.
\end{itemize}

While the quantifiable impact of these design choices has yet to be formally evaluated, we intend to measure and present these metrics in future work.

In BT-based behaviours, BTs utilise the ticking function along with the return value from each subtree to trigger the next candidate. In FSM-based behaviours, FSMs use the event system and information from the previous state to determine the next state. As a result of the previous design decision, the trigger mechanism needed to be compatible with part of the Behaviour sequence being invisible/inactive. This aspect of BTs makes it less suitable for long-horizon tasks. Since BTs trigger the entire tree, the entire BT must be in the computer's memory, and thus, our requirements cannot be met with BTs. A more general graph (PSFSM) structure allows for unknowns by design. Due to this reason, we selected FSM as our basis.

We selected the \textbf{Capabilities2} system as our base because it supports loading and unloading its \textit{Runners} (its basic building block) from memory \cite{pritchard_capabilities2_2025}. Also, its \textit{Interface} and \textit{Provider} files fulfil the first requirement. We augmented the \textbf{Capabilities2} framework with an \textit{Event subsystem} and a \textit{parameter-based instantiation} to emulate the behaviour of an FSM and fulfil the fourth requirement.

To fulfil the second requirement, we propose \textbf{Fabric}, a system that supports \textit{parsing}, \textit{validation}, and \textit{conversion} of a \textit{Behaviour Plan} into \textbf{Capabilities2} compatible instructions. 

To fulfil the third requirement, we present \textit{PromptTools}, a ROS2 package that interfaces LLMs with the ROS2 ecosystem via a standard ROS2 service interface, and \textit{Runners} for \textbf{Fabric} and \textbf{Capabilities2}, which can be used for generative functionality. Fig. \ref{fig-system} shows the system structure of the three subsystems.

In the following sections, we present \textbf{Fabric} and modifications to \textbf{Capabilities2} and \textbf{PromptTools}, respectively.
    
\subsection{Fabric}

\begin{figure*}[t]
\vspace{1ex} 
\centering
\hfill
\begin{subfigure}[t]{0.47\textwidth}
    \centering
    \includegraphics[width=\linewidth]{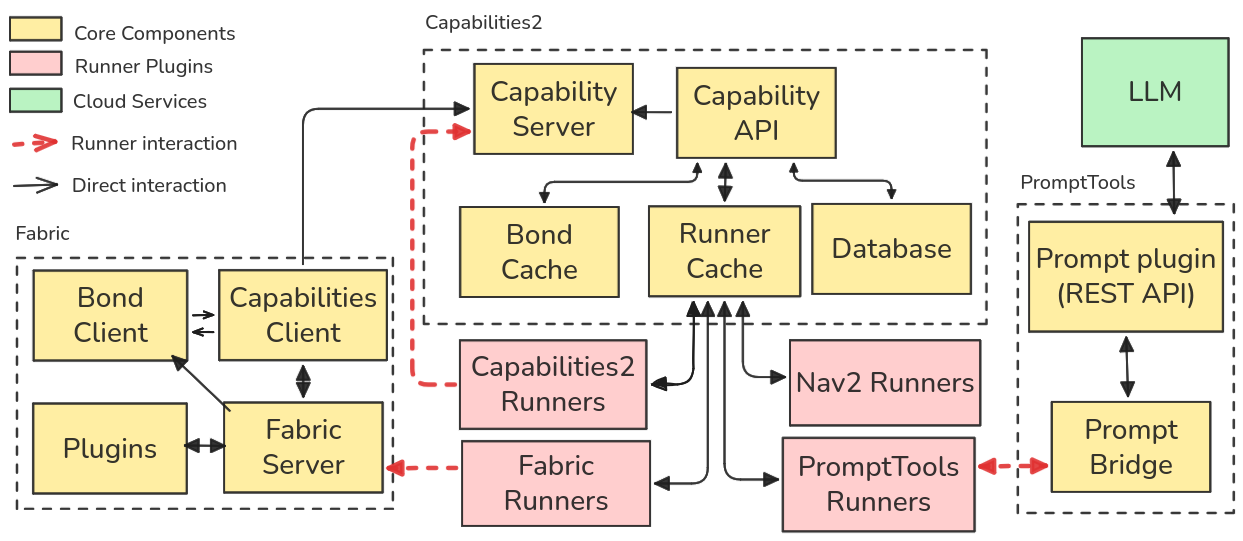}
    \caption{Structure}
    \label{fig-system}
\end{subfigure}
\hfill
\begin{subfigure}[t]{0.45\textwidth}
    \centering
    \includegraphics[width=\linewidth]{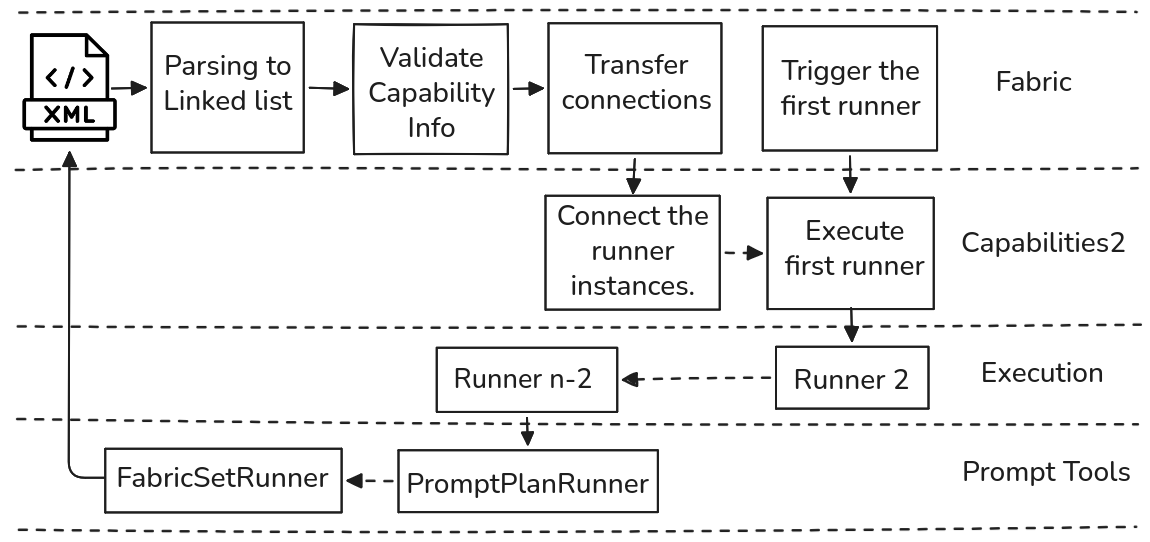}
    \caption{Generative Process pipeline}
    \label{fig-generation-pipeline}
\end{subfigure}
\hfill
\caption{Structure and Generative Process pipeline of the Fabric, Capabilities2 and PromptTools}
\label{fig-system-and-pipeline}
\end{figure*}

As shown in Fig. \ref{fig-system}, the \textbf{Fabric} comprises 3 components: a \textit{Fabric Server}, a \textit{Capabilities2 Client}, and a \textit{Bond Client}. The server initialises by loading parsing and validation plugins and loading the initial execution plan from a file. The parser plugin reads the execution plan and extracts connections between behaviours. The capabilities2 client retrieves data from the \textbf{Capabilities2} system, and the validation plugin uses that data to validate the plan's compatibility with the robot. Once validated, the \textit{Capabilities2 client} transfers the connection information to the \textit{Capabilities2 event subsystem}. The \textit{Bond Client} maintains a heartbeat signal with the \textbf{Capabilities2} system as a failsafe in case the \textbf{Fabric} system fails. Figure \ref{fig-generation-pipeline} showcases this process.

\subsubsection{Execution plan} 
\begin{figure*}[htbp]
\hfill
\centering
\begin{subfigure}[t]{0.33\textwidth}
  \centering
  \includegraphics[width=\linewidth]{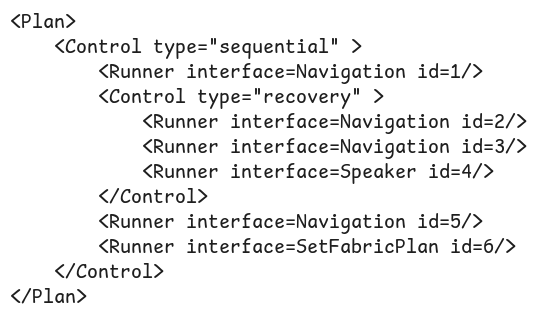}
  \caption{A simplified XML-based behaviour plan}
  \label{fig-plan-and-fabric-a}
\end{subfigure}
\hfill
\begin{subfigure}[t]{0.33\textwidth}
  \centering
  \includegraphics[width=\linewidth]{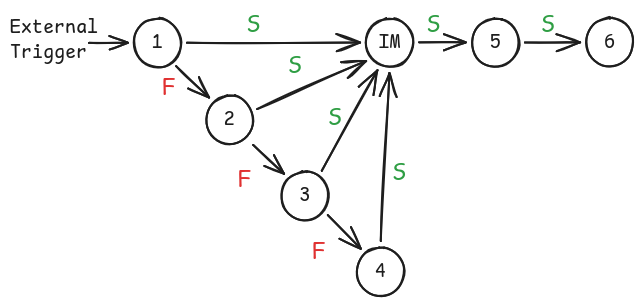}
  \caption{The graphical representation of the plan}
  \label{fig-plan-and-fabric-b}
\end{subfigure}
\hfill
\begin{subfigure}[t]{0.27\textwidth}
  \centering
  \includegraphics[width=\linewidth]{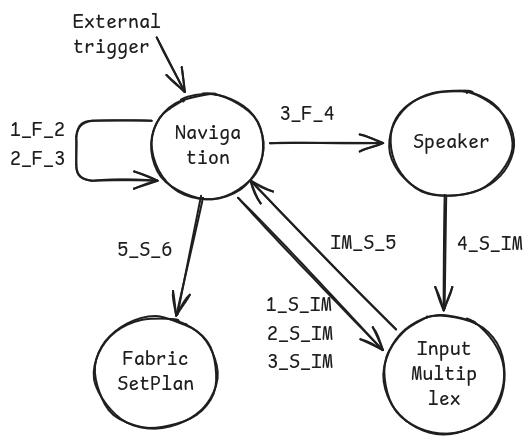}
  \caption{Runners executed in the system. }
  \label{fig-plan-and-fabric-c}
\end{subfigure}
\hfill
\caption{XML-based execution plan and its Fabric Instance. "S" and "F" represent successful and failed events. 1\_S\_2 means a success event from runner instance 1 to runner instance 2.}
\label{fig-plan-and-fabric}
\end{figure*}

describes the execution flow of behaviours. Fig. \ref{fig-plan-and-fabric} contains an example. The Execution plan supports two types of XML elements: \textit{Control elements} and \textit{Runner elements}. A \textit{Runner element} implements a behaviour or part of a behaviour. During execution, \textit{Runners} in the system are unique and start only once, and multiple instances are handled through threads. \textit{Control elements} dictate how the \textit{Runner elements} would be connected to each other. Currently, the \textbf{Fabric} supports the following control elements,

\begin{figure*}[t]
\hfill
\begin{subfigure}[t]{0.22\textwidth}
  \centering
  \includegraphics[width=\linewidth]{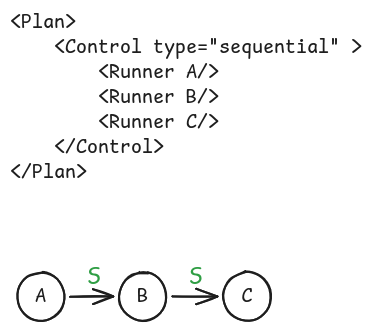}
  \caption{Sequential control flow}
  \label{fig-control-blocks-a}
\end{subfigure}
\hfill
\begin{subfigure}[t]{0.22\textwidth}
  \centering
  \includegraphics[width=\linewidth]{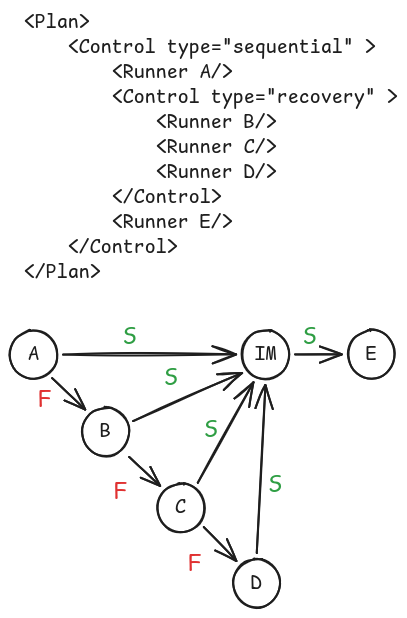}
  \caption{Recovery control flow}
  \label{fig-control-blocks-b}
\end{subfigure}
\hfill
\begin{subfigure}[t]{0.22\textwidth}
  \centering
  \includegraphics[width=\linewidth]{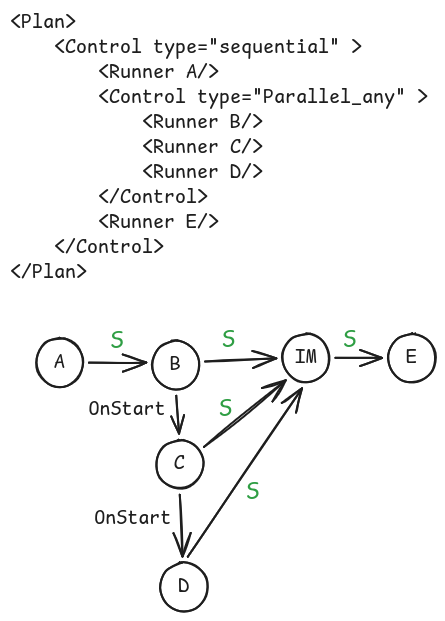}
  \caption{Parallel any/all control flow}
  \label{fig-control-blocks-c}
\end{subfigure}
\hfill
\caption{Control structures used. "S" and "F" represent successful and failed events.  Onstart represents an "OnStarted" Event.}
\label{fig-control-blocks}
\end{figure*}

\begin{itemize}
    \item \textit{Sequential} control executes the child runners sequentially upon the successful completion of their immediate predecessors. Shown in figure \ref{fig-control-blocks-a}.
    \item \textit{Recovery} control executes child runners only when the immediate predecessor outside the control block fails. Children within the control block are triggered until one succeeds; at that point, the control block exits and triggers the successor block/runner, ignoring the other children in the block. Shown in figure \ref{fig-control-blocks-b}
    \item \textit{Parallel Any} control executes all child runners at once and waits until at least one completes before triggering the successor. Shown in figure \ref{fig-control-blocks-c}.
    \item \textit{Parallel All} control executes child runners simultaneously and waits for all runners to complete before triggering the successor runner/block.  Shown in figure \ref{fig-control-blocks-c}.
\end{itemize}

These control blocks dictate how the runners (and thus behaviours) are connected to each other, and to implement \textit{Recovery}, \textit{Parallel Any} and \textit{Parallel All} functionalities, we add internal \textit{Runners} such as \textit{IM (InputMultiplex)} as in Figure \ref{fig-control-blocks} to handle multiple trigger commands.

\subsubsection{Fabric capabilities} allows the created Fabric to manipulate the \textit{Fabric server} itself during runtime. Currently, it supports runners that load a new plan onto the \textit{Fabric server} and notify Fabric's own interface when the current plan completes, as shown in figure \ref{fig-generation-pipeline}. This enables dynamic loading of the next execution plan while the current plan is executing.
    
\subsection{Capabilities2}

A behaviour can either be a single \textit{Runner} or a \textit{composition of Runners}. A \textit{Runner} can be a ROS 2 action client, a service client, or a topic subscriber, among other complex implementations. \textit{A Runner represents a node in the GPSFSM graph}. The main modification to the \textbf{Capabilities2} system includes,

\subsubsection{Start-Trigger separation} in which we modified the original \textit{Start} functionality in the runner implementation into \textit{Start} and \textit{Trigger} functionalities, separating the starting process (loading the capability into the system along with dependencies) from the triggering process (executing). We also added parameter-based instantiation for the \textit{Runner} during runtime, allowing LLMs (and runners) to configure and modify runners (and sibling runners).

\subsubsection{Event subsystem} allows asynchronous event-based triggering and supports four event types \textit{STARTED}, \textit{STOPPED}, \textit{SUCCESS}, and \textit{FAILED}. This allows \textit{tracking the runner's state and linking runners to others, enabling triggering a child based on the parent runner's state}. To maintain the original design goals, we added thread-based execution and event-layering, enabling us to run the same runner across multiple events without increasing resource utilisation. \textit{Event connections represent edges in the GPSFSM graph}. \textit{Unconnected Events are either ignored as 'don't care' states or collected in fallback structures, such as 'give up', which then leads to 'regenerate'}.

\subsubsection{Capabilities2 Runners} allows the \textbf{Fabric} to interact with the \textit{Capability server} via \textit{Runners} and also contains \textit{Internal Runners} to implement \textit{Recovery}, \textit{Parallel Any}, and \textit{Parallel All}. For \textit{Recovery} and \textit{Parallel Any} sequences, the \textit{Internal Runner} waits for at least one input before forwarding the trigger, while \textit{Parallel All} waits for all defined trigger inputs before forwarding the trigger.

Fabric utilises this separation of \textit{Start-Trigger functionality to configure the runners into an FSM} and the \textit{Event subsystem to initiate a cascading triggering effect over time}. Complex behaviour executions would propagate like ripples through a fabric.
    
\subsection{PromptTools}

PromptTools is a ROS 2 package that provides a unified RESTful service for accessing local and cloud LLMs. It uses a plugin-based architecture, allowing model provider plugins to be swapped at runtime based on the requested service, with current plugins supporting Ollama and OpenAI GPT models.

\subsubsection{Prompt buffering} enables the robot to accumulate prompt fragments incrementally and flush them on demand. textit{Runners} use this to independently contribute information to the buffer, and when the buffer is flushed, PromptTools aggregates their inputs into a single, coherent prompt before issuing the final prompt with the desired task and/or context.

\subsubsection{Prompt Runners} allow for triggering the PromptTools system with new information as the Fabric progresses. Currently, there are \textit{Runners} for prompting Nav2-related information, capability information, new task assignment, and new execution plan generation.

\subsection{Generative functionality}

These three subsystems, \textbf{Fabric}, \textbf{Capabilities2}, and \textbf{PromptTools}, along with their runner plugins, enable Fabric to function as both a generic FSM and a Generative FSM within the same codebase. In the generic functionality, the execution plan can include any combination of Control and Runner Elements, and only the \textbf{Fabric} and \textbf{Capabilities2} subsystems are required. To achieve generative functionality, a process pipelining similar to that shown in the figure. \ref{fig-generation-pipeline} is necessary to initialise the task generation process. Once started, the execution plans and GPSFSM instances continue to generate until the task is complete.

%%%%%%%%%%%%%%%%%%%%%%%%%%%%%%%%%%%%%%%%%%%%%%%%%%%%%%%%%%
%%%% Results
%%%%%%%%%%%%%%%%%%%%%%%%%%%%%%%%%%%%%%%%%%%%%%%%%%%%%%%%%%

\section{Experimental Results and Discussion}

When evaluating the generative qualities of an FSM or a BT, the quality of the generated execution plan \cite{izzo_btgenbot_2024, lykov_llm-brain_2023} has been a key evaluation metric, along with factors such as time consumption. We conduct experiments as explained in the following sections. The first experiment focuses on generative quality, and the second experiment focuses on complex behaviours.

\begin{figure}[htbp]
\centering

\begin{subfigure}[t]{0.58\columnwidth}
  \centering
  \includegraphics[width=\linewidth]{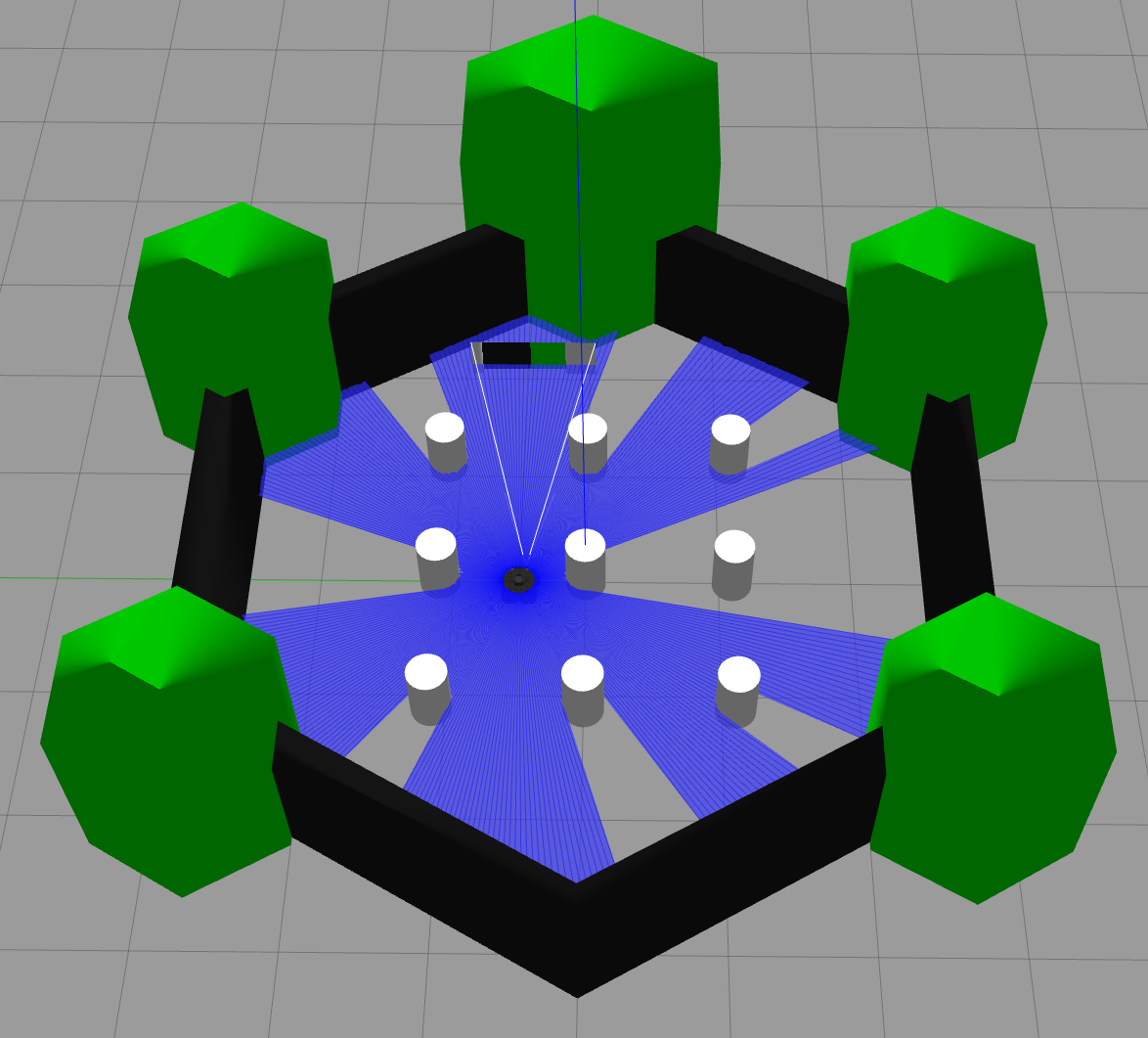}
  \caption{Simulation Environment}
  \label{fig-simenv}
\end{subfigure}\hfill
\begin{subfigure}[t]{0.40\columnwidth}
  \centering
  \includegraphics[width=\linewidth]{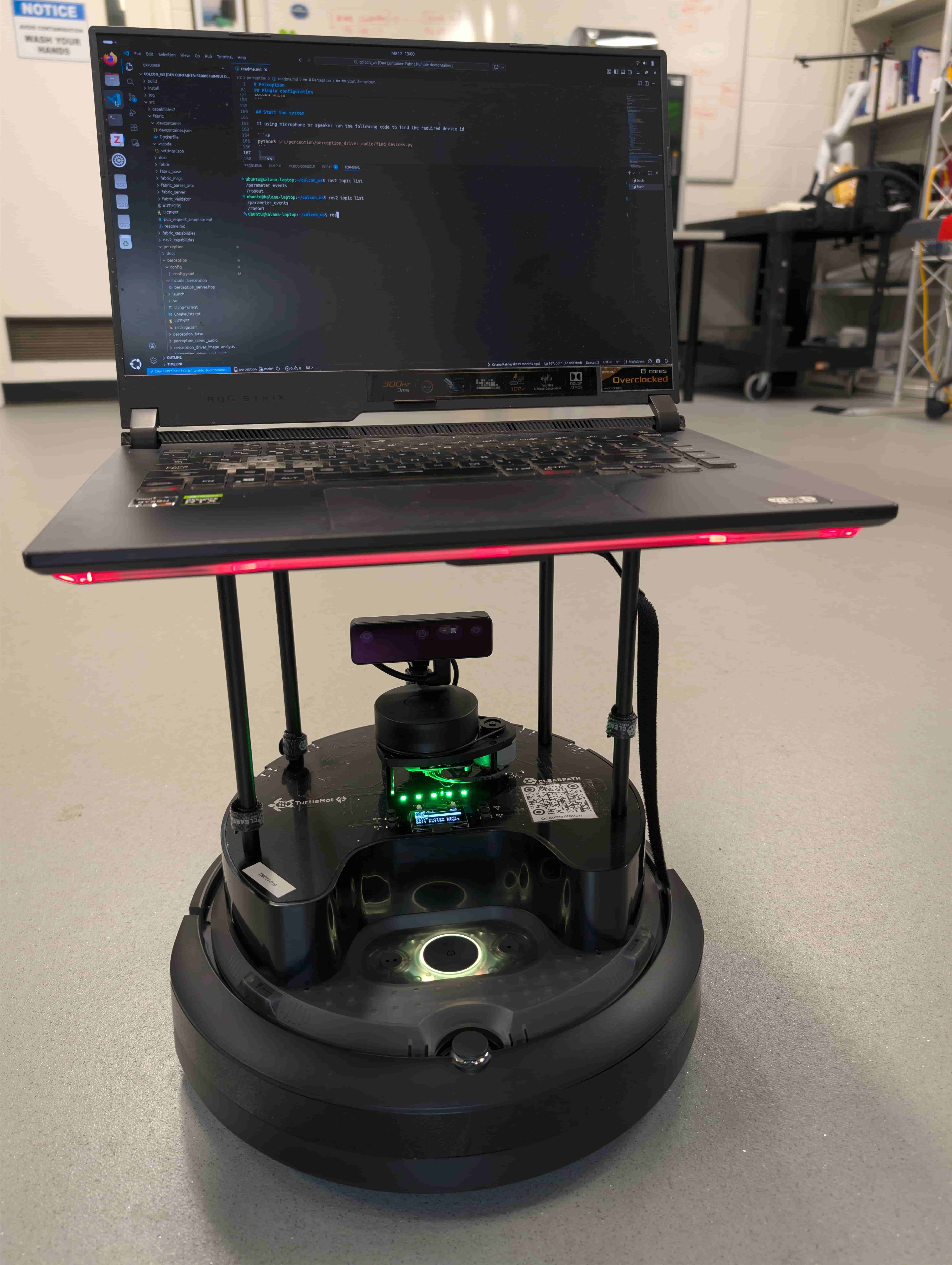}
  \caption{Physical robot}
  \label{fig-physical}
\end{subfigure}

\caption{Testing environments}
\label{fig-testing}
\end{figure}

\subsection{Generative Experiment}

In the first experiment, we evaluate generative functionality. Our experiments are based on navigation tasks because they are easily reproducible in both simulation and simple hardware with minimal changes. 

We evaluate our system on five navigation tasks in simulation (see figure \ref{fig-simenv}), with increasing difficulty, focusing on waypoint navigation and gradually introducing unreachable points and their alternatives. The evaluation tasks include,

\begin{itemize}
    \item \textit{Task 1:} single navigation point
    \item \textit{Task 2:} four navigation points
    \item \textit{Task 3:} four navigation points with one recovery point
    \item \textit{Task 4:} four navigation points with two recovery points
    \item \textit{Task 5:} four navigation points with four recovery points
\end{itemize}

Tasks 1 and 2 represent normal behaviour sequences. For tasks 3, 4, and 5, recovery behaviours are integrated with normal behaviours, with some expected to be utilised and others not required.

We utilise local and cloud LLMs in our evaluations to study their impact on our system. Local LLMs are run on a computer equipped with a 16-core Intel Xeon CPU, 256 GB of RAM, and an NVIDIA RTX 3090 GPU with 24GB of memory. We use the same network to connect with GPT models and measure generation time from request to response as a key evaluation metric. The timing measures are averaged and available in the Table \ref{Table-Eval-Overall}.

\subsubsection{Benchmarking}

BTGenBot \cite{izzo_btgenbot_2024} is the SOTA that provides LLM-based execution plan generation along with execution (though on independent pipelines without dynamic BT loading). It utilises local LLMs for BT generation and utilises BehaviourTree.CPP \cite{ghzouli_behavior_2023} as the BT engine. Additionally, it supports the ROS2 ecosystem, building on existing support for navigation and manipulation. For these reasons, we benchmark our system against BTGenBot.

To benchmark generative quality, we extract the generative pipelines from both systems and evaluate them against the local and cloud LLMs used by each system. This includes locally run LLamaChat and Codellama models, as well as OpenAI GPT-4o, 4.1, and 5 models. To balance the scales:

\begin{itemize}
    \item We limit the prompt used for Fabric-related testing to include only information on Nav2 waypoint navigation and plan generation.
    \item We modify the oneshot prompt of BTGenBot to include recovery functionality, since the original did not include examples of recovery functions.
    \item We utilise the original model base and the fine-tuned version provided by the BTGenBot authors for BT generation, without additional fine-tuning for Fabric plans.
\end{itemize}

We run the generative pipelines for both systems, with 10 iterations per task in zero-shot and one-shot prompt modes, across 7 LLM types. During the evaluation process, plans were tested in simulation alongside three human evaluators, who assessed their syntactic and semantic accuracy and categorised them as Failed, Partial, or Successful. Failed categories include corrupted plans and incomplete plans. Partial categories include plans that are either syntactically or semantically correct but not both. Successful plans are both syntactically and semantically correct. Table \ref{Table-Eval-Overall} summarises the evaluation results for each task, and the detailed analysis is available in the repository. Data from tests 1-5 are aggregated using the following equations. For each row and framework, let $i$ index the non-corrupted tasks, with $n_i$ valid runs (here typically $10$), mean time $\mu_i$, and sample standard deviation $\sigma_i$.

\begin{equation}
S=\sum_i S_i, \quad P=\sum_i P_i, \quad F=\sum_i F_i, \quad N=\sum_i n_i
\end{equation}
\begin{equation}
\mu=\frac{\sum_i n_i\mu_i}{N}
\end{equation}
\begin{equation}
\sigma=\sqrt{\frac{\sum_i (n_i-1)\sigma_i^2 + \sum_i n_i(\mu_i-\mu)^2}{N-1} }
\end{equation}

\subsubsection{Results}

% -----------------------------------------------------------------------------
% Combined summary (Tasks 1--5)
% -----------------------------------------------------------------------------
\begin{table*}[t]
\centering
\vspace{2ex} 
\caption{Overall summary across Tasks 1--5 (corrupted results excluded).}
\label{Table-Eval-Overall}
\vspace{1ex} 
    \begin{tabular}{|c|c|c|cc|cc|}
    \hline
    \textbf{Model} & \textbf{Version} & \textbf{Prompt} & \multicolumn{2}{c|}{\textbf{Fabric}} & \multicolumn{2}{c|}{\textbf{BTGenBot}} \\
    \cline{4-7}
    & & & \textbf{S/P/F (N)} & \textbf{Time $\mu\pm\sigma$ (s)} & \textbf{S/P/Fixed (N)} & \textbf{Time $\mu\pm\sigma$ (s)} \\
    \hline
    Codellama & Base      & Zeroshot & 20/30/0 (N=50) & 23.04 $\pm$ 9.21 & 0/20/30 (N=50) & 26.44 $\pm$ 13.55 \\
              & Base      & Oneshot  & 0/0/0 (N=0)    & ---              & 10/0/30 (N=40) & 27.88 $\pm$ 17.55 \\
              & Finetuned & Zeroshot & 20/30/0 (N=50) & 23.01 $\pm$ 9.25 & 0/20/30 (N=50) & 26.33 $\pm$ 13.29 \\
              & Finetuned & Oneshot  & 0/0/0 (N=0)    & ---              & 10/0/30 (N=40) & 28.04 $\pm$ 17.56 \\
    \hline
    Llamachat & Base      & Zeroshot & 0/16/34 (N=50) & 39.89 $\pm$ 48.49 & 0/17/33 (N=50) & 23.94 $\pm$ 20.33 \\
              & Base      & Oneshot  & 0/0/50 (N=50)  & 0.89 $\pm$ 1.59  & 17/27/6 (N=50) & 16.39 $\pm$ 12.46 \\
              & Finetuned & Zeroshot & 0/13/37 (N=50) & 31.80 $\pm$ 24.31 & 0/16/34 (N=50) & 24.81 $\pm$ 17.40 \\
              & Finetuned & Oneshot  & 0/0/50 (N=50)  & 0.87 $\pm$ 1.67  & 14/30/6 (N=50) & 18.29 $\pm$ 13.44 \\
    \hline
    GPT & 4o  & Zeroshot & 31/19/0 (N=50) & 2.64 $\pm$ 1.11 & 0/0/50 (N=50) & 1.73 $\pm$ 0.65 \\
        & 4o  & Oneshot  & 49/1/0 (N=50)  & 2.51 $\pm$ 0.88 & 50/0/0 (N=50) & 2.12 $\pm$ 0.62 \\
        & 4.1 & Zeroshot & 45/5/0 (N=50)  & 2.33 $\pm$ 1.04 & 0/8/42 (N=50) & 1.54 $\pm$ 0.47 \\
        & 4.1 & Oneshot  & 46/3/0 (N=49)  & 2.16 $\pm$ 0.73 & 50/0/0 (N=50) & 1.46 $\pm$ 0.63 \\
        & 5   & Zeroshot & 50/0/0 (N=50)  & 19.94 $\pm$ 8.83 & 13/25/12 (N=50) & 15.21 $\pm$ 5.47 \\
        & 5   & Oneshot  & 50/0/0 (N=50)  & 16.15 $\pm$ 8.47 & 50/0/0 (N=50) & 9.66 $\pm$ 3.67 \\
    \hline
    \end{tabular}
    \vspace{1ex} 
    {\centerline{\footnotesize S/P/F are aggregated counts over valid runs only; $N$ can vary due to excluding corrupted ($^{\mathrm{x}}$) rows.}}
\end{table*}

The most significant result, as shown in Table \ref{Table-Eval-Overall}, is that Fabric plan generation for all tasks across all GPT models, with zero-shot and one-shot settings, performs exceptionally well. (90\% success, 10\% partial) In comparison, BTGenBot achieves a relatively low success rate (54\% successful, 11\% partially successful, and 34\% failed). The success of BTGenBot is primarily attributed to one-shot prompting, suggesting that failures may stem from a lack of examples in the prompt. Retrospectively, this highlights that the modular prompting approach, with runner-plugin-based descriptions and XML elements used in Fabric, is successful.

In local models, BTGenBot performs better (13\% success, 35\% partial, 52\% failed) than Fabric (10\% success, 22\% partial, 68\% failed). In response to the one-shot prompt for fabric, both codellama and llamachat returned the prompt example. These responses are indicated as corrupted using $^x$ in the Table \ref{Table-Eval-Overall}. This could be because the prompt is too large for the LLM (i.e., the context window is too small). The fact that local base LLMs (both Codellama and Llamachat) succeed (40\% success and 60\% partial) in generating a fabric execution plan via zero-shot prompting (which requires shorter context) further reinforces this. Most failures of the BTGenBot and Fabric occur in more complex tasks. Failures in the generative stage include invalid leaf node names, parameters, or tree structures for BTGenBot, and invalid runner names, parameters, or runner nesting errors for Fabric.

Comparing generation times indicates that the overall increase in task complexity is accompanied by a corresponding increase in generation time. There is a considerable difference between cloud LLMs and Local LLMs in the generation time for both Fabric and BTGenBot. Outliers in this regard are the GPT-5 models, which deviate from this pattern and remain in the middle. This could be due to the GPT-5 model's high reasoning capabilities, as evidenced by its successful plan generation for BTGenBot zero-shot generation, whereas GPT-4.1 and GPT-4o failed.

Experiment setting up, prompts, and resulting XML plans are available at \textit{https://github.com/CollaborativeRoboticsLab/gpsfsm-2026}

\subsection{Complex Behaviour Experiment}

In the second experiment, we test complex behaviours. These experiments are based on navigation and perception (audio, video, transcription, image analysis) tasks because they can be integrated with complex behaviours.

We demonstrate our system on 3 complex real-world tasks, increasing in difficulty and focusing on waypoint navigation and human interaction, gradually introducing increasingly complex commands that involve speech and audio feedback.

\begin{itemize}
    \item \textit{Task 1:} moving to 1 meter and describing surroundings
    \item \textit{Task 2:} moving to a given location, asking for a person's name, returning, and repeating the name.
    \item \textit{Task 3:} moving to a given location, asking for a person's name, returning and repeating the name. If the person is not there, move to a different location.
\end{itemize}

We utilise Cloud LLMs for this, along with a Turtlebot4 robot with a laptop on top. The robot runs the localisation and nav2 stacks while the laptop runs the perception, fabric, capabilities2 and prompt tools stacks.

\textit{Results:} In this experiment, we do not evaluate against a specific system because we could not find an equivalent system that implements the full generation pipeline. Instead, we utilise these experiments to showcase our systems' capabilities. Experiment videos, logs, and generated execution plans are available at \textit{https://github.com/CollaborativeRoboticsLab/gpsfsm-2026}

%%%%%%%%%%%%%%%%%%%%%%%%%%%%%%%%%%%%%%%%%%%%%%%%%%%%%%%%%%
%%%% Future Work
%%%%%%%%%%%%%%%%%%%%%%%%%%%%%%%%%%%%%%%%%%%%%%%%%%%%%%%%%%

\section{Conclusion and Future Work}

We presented the first systematic approach to Generative Partially Specified Finite State Machines for robotic behaviour planning, demonstrating that FSMs, when coupled with the Capabilities2 system, provide smooth behaviour generation and transitions via event-triggered transitions and LLM integration compared to hierarchical BTs. Our comprehensive ROS2 stack (Fabric, Capabilities2, and PromptTools) enables practical GPSFSM deployment while introducing standardised semantic capability descriptions that are currently lacking in generative systems.

Experimental evaluation validates our semantic complexity hypothesis: our GPSFSM approach achieved higher success with GPT models in both zero-shot and one-shot scenarios than the state-of-the-art BT-based generation from BTGenBot. This performance gap, particularly in zero-shot generation, confirms that Fabric's simpler state-based semantics facilitate more reliable LLM reasoning without extensive prompt engineering. However, local LLMs struggled with context window limitations, revealing a trade-off between comprehensive capability descriptions and resource-constrained deployments that warrants further investigation.

In future work, we plan to extend our system to support robot arm manipulation, thereby broadening the range of robotic platforms it can operate on and enabling more comprehensive evaluations. We also aim to benchmark our approach against Behaviour Trees (BTs) to assess computational efficiency and runtime overhead. Beyond these directions, we see opportunities to enhance the system by incorporating execution-plan fault detection and correction using LLM-based multi-agent reasoning before dispatching plans to Fabric, and by implementing environmental-interrupt handling through robotic attention mechanisms. These extensions will further strengthen the robustness, adaptability, and practical applicability of our framework.

%%%%%%%%%%%%%%%%%%%%%%%%%%%%%%%%%%%%%%%%%%%%%%%%%%%%%%%%%%
%%%% Acknowledgment
%%%%%%%%%%%%%%%%%%%%%%%%%%%%%%%%%%%%%%%%%%%%%%%%%%%%%%%%%%

\section*{Acknowledgment}

This research was supported by an Australian Government Research Training Program (RTP) Scholarship \url{https://doi.org/10.82133/C42F-K220}

%%%%%%%%%%%%%%%%%%%%%%%%%%%%%%%%%%%%%%%%%%%%%%%%%%%%%%%%%%
%%%% References
%%%%%%%%%%%%%%%%%%%%%%%%%%%%%%%%%%%%%%%%%%%%%%%%%%%%%%%%%%

\raggedbottom
\bibliographystyle{ieeetr}
\bibliography{references}

\end{document}